\documentclass[twocolumn,a4paper]{article}
\usepackage{graphicx}
\usepackage{amsmath, amssymb}
\usepackage{hyperref}
\usepackage{booktabs}
\usepackage{subcaption}
\usepackage{float}
\usepackage{todonotes}
\usepackage{longtable}
\usepackage{subcaption,graphicx}
\usepackage{dblfloatfix}
\usetikzlibrary{shapes,arrows,positioning,fit,backgrounds}

\usepackage[backend=biber]{biblatex}
\title{It's 2025 -- Narrative Learning is the new baseline to beat for explainable machine learning}
\author{Gregory D. Baker \\ Australian National University \\ greg.baker@anu.edu.au}

\usepackage{booktabs,tabularx,makecell,siunitx}
\begin{document}
\maketitle

\tikzset{
    block/.style={rectangle, draw, fill=blue!10, text width=6em, text centered, rounded corners, minimum height=3em},
    data/.style={rectangle, draw, fill=green!10, text width=5em, text centered, minimum height=2.5em},
    model/.style={ellipse, draw, fill=orange!20, text width=5em, text centered, minimum height=3em},
    arrow/.style={thick,->,>=stealth},
    dashedarrow/.style={thick,->,>=stealth,dashed}
}

\begin{figure}[b!]
  \scalebox{0.6}{
\begin{tikzpicture}[node distance=1cm, auto]

\node [data] (train) {Training Data};

\node [model, below=of train] (overseer) {Overseer LLM};

\node [block, below=of overseer] (prompt) {Narrative Prompt};

\node [model, below=of prompt] (underling) {Underling LLM};

\node [data, left=1.5cm of underling] (val) {Validation Data};

\node [block, right=1.5cm of underling] (metrics) {Performance Metrics};

\node [block, below=1.5cm of underling] (ensemble) {Ensemble\\(3 models)};

\node [data, left=1.5cm of ensemble] (test) {Test Data};

\node [data, below=of ensemble] (output) {Predicted Labels};

\draw [arrow] (train) -- node[right,font=\footnotesize] {examples} (overseer);
\draw [arrow] (overseer) -- node[right,font=\footnotesize] {generates} (prompt);
\draw [arrow] (prompt) -- (underling);
\draw [arrow] (underling) -- (ensemble);
\draw [arrow] (ensemble) -- (output);

\draw [arrow] (val) -- (underling);
\draw [arrow] (underling) -- node[above,font=\footnotesize] {evaluate} (metrics);

\draw [dashedarrow] (metrics) |- node[right,font=\footnotesize] {feedback} (overseer);

\draw [arrow] (test) -- (ensemble);


\end{tikzpicture}
}
\caption{Narrative Learning flow diagram}
\label{flow-diagram}
\end{figure}
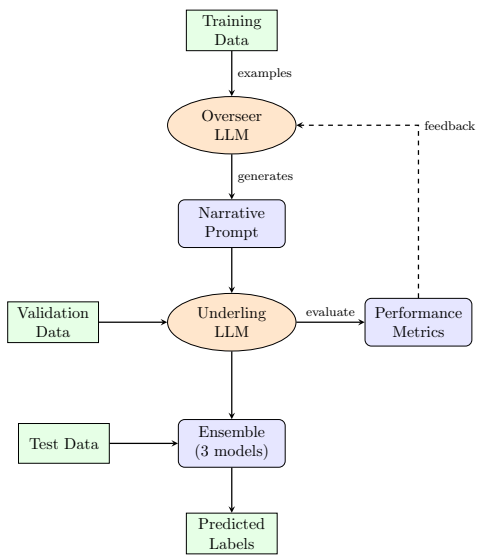

\begin{abstract}
  In this paper, we introduce Narrative Learning, a methodology where
  models are defined entirely in natural language and iteratively
  refine their classification criteria using explanatory prompts
  rather than traditional numerical optimisation. We report on
  experiments to evaluate the accuracy and
  potential of this approach using 3 synthetic and 3 natural datasets
  and compare them against 7 baseline explainable machine learning
  models.  We demonstrate that on 5 out of 6 of these datasets,
  Narrative Learning became more accurate than the baseline
  explainable models in 2025 or earlier because of improvements in
  language models.  We also
  report on trends in the lexicostatistics of these models' outputs as
  a proxy for the comprehensibility of the explanations.
\end{abstract}

\section{Motivation}
Machine learning models are often categorised into ``explainable models''
(examples include logistic regression\cite{Cox1958},
decision trees\cite{Breiman1984},
Bayesian rule lists\cite{Letham2015},
CORELS\cite{Angelino2018},
explainable boosting machines\cite{Nori2019}
and
rulefit\cite{Friedman2008}) and ``blackbox models''
(most neural network architectures fit into this category). A model
is considered ``explainable'' if there is a human language set of instructions
that can be followed by a human being that produces the same result as
the computational model; hopefully the process of producing these same
results is somehow enlightening to the human being in understanding the
interactions of the features of the model to produce the result.

This is not the same as an explainability layer over a blackbox model
--- popular techniques including feature attribution methods such as
LIME and SHAP \cite{Ribeiro2016, Lundberg2017} --- where the model itself is inscrutable
to human understanding, but a post-training step creates some insight
for the human being.

An explanation is considered concordant with the model if it produces
the same result under all circumstances: an explainable model will do
this perfectly, and an explainability layer will only give an
approximation to being concordant. There are legal requirements in
some jurisdictions (e.g. EU GDPR\cite{GDPR} and China's
PIPL\cite{PIPL}) where better explainability can often be more
valuable than better accuracy.  It can be worthwhile to pay an
additional computational cost and accept the lower accuracy of using a
fully-explainable model in exchange for the legal guarantee that the
explanation provided to the customer is precisely the model that was
used for producing the result.

Traditionally, the human-language instructions and the computational
model for the explainable model have been separate objects, but with recent
advancements in large language models (LLMs), AI systems can now
comprehend and execute human-readable explanations, allowing for
the possibility of having just a single human-language model.

Thus the dual modelling (computational form for the computer,
explainable form for the human) can be replaced by a unified model (an
explainable form that is read and executed by computer and human
alike). We name this approach \textit{Narrative Learning} --- the narrative
that is provided to the human being is the object that needs to be learned
by the machine learning process.

The narrative that is learned can take any form, and we observe
that the LLMs create quite a variety of different outputs, as
shown in Appendix \ref{rules-generated}. Note that those examples
(and the example in Figure~\ref{titanic-sample}) are the reverse
translated versions of the actual instructions. We choose not to provide
the actual instructions (on the transformed datasets that we used) in
order to keep them out of the training set in future language models.

\begin{figure}[t!]
{\sf \footnotesize
  You are given details about a single passenger: Pclass, Sex, Age, SibSp, Parch, Fare, and Embarked. Apply these rules in order:

\begin{enumerate}
\item If SibSp $\ge$ 4, label the outcome as 0.
\item If the passenger is male and Embarked is ``S'' and Age $\ge$ 3.33, label the outcome as 0.
\item If the passenger is female and Embarked is ``Q'' and Fare $<$ 7.7, label the outcome as 0.
\item Otherwise, label the outcome as 1.
\end{enumerate}

Provide only the single digit "1" or "0" as your output.
}
\caption{Reverse translated Narrative Learning output for the
  Titanic dataset (OpenAI o1 with 10 examples)}
\label{titanic-sample}
\end{figure}

\begin{table*}[b!]
  \begin{center}
  \begin{tabular}{lrrr}
    {\bf Dataset} & {\bf Annual improvement} & {\bf Trend p-value} & {\bf SOTA Date} \\
    Titanic & 0.217 & 0.0051 & 2025-01-27 \\
    Wisconsin & 0.271 & 0.0036 & 2025-02-21 \\
    SG Credit &  -0.043 & 0.18 & N/A \\
    Espionage &  0.231 & 0.066 & 2024-11-29 \\
    Time Travel & 0.065 & 0 & Before 2024 \\
    Potions &  0.056 & N/A & 2024-05-09 \\
  \end{tabular}
  \end{center}
  \caption{Narrative Learning improvement rates in negative log KT
    accuracy, trend p-value (where there are more than two points) and date when it is predicted to be the state-of-the-art
    explainable machine learning algorithm, or the date when it became the state of the art}
  \label{pvalues-and-trends}
\end{table*}

\begin{figure*}[t!] 
  \centering
  \begin{subfigure}[t]{0.49\textwidth}
    \centering
    \includegraphics[width=\linewidth,trim=2mm 2mm 2mm 2mm,clip]{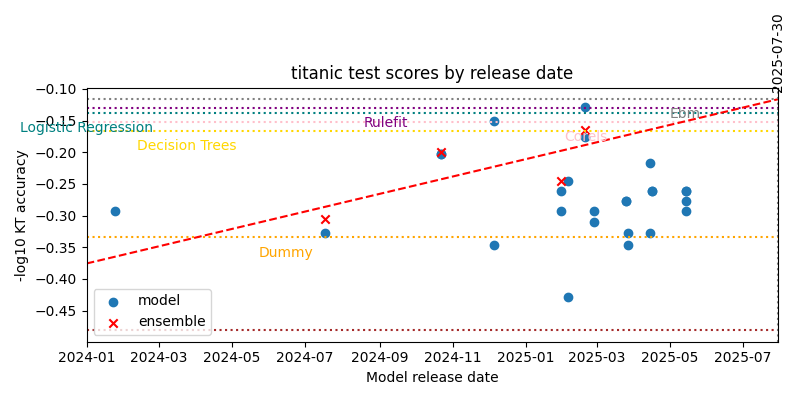}
    \caption{Titanic}\label{fig:titanic}
  \end{subfigure}\hfill
  \begin{subfigure}[t]{0.49\textwidth}
    \centering
    \includegraphics[width=\linewidth,trim=2mm 2mm 2mm 2mm,clip]{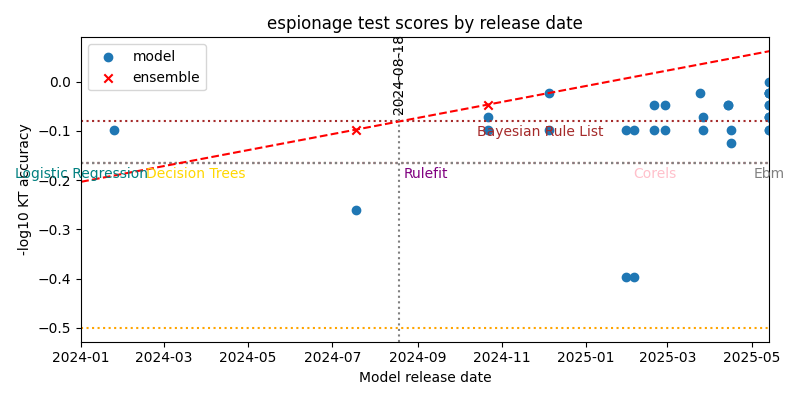}
    \caption{Espionage}\label{fig:espionage}
  \end{subfigure}

  \medskip
  \begin{subfigure}[t]{0.49\textwidth}
    \centering
    \includegraphics[width=\linewidth,trim=2mm 2mm 2mm 2mm,clip]{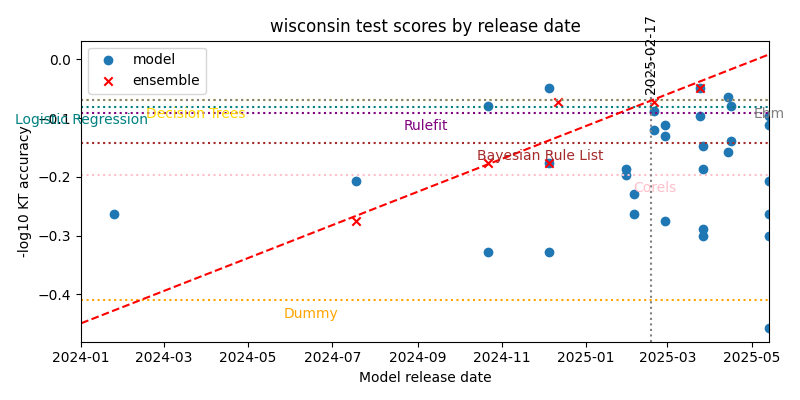}
    \caption{Wisconsin}\label{fig:wisconsin}
  \end{subfigure}\hfill
  \begin{subfigure}[t]{0.49\textwidth}
    \centering
    \includegraphics[width=\linewidth,trim=2mm 2mm 2mm 2mm,clip]{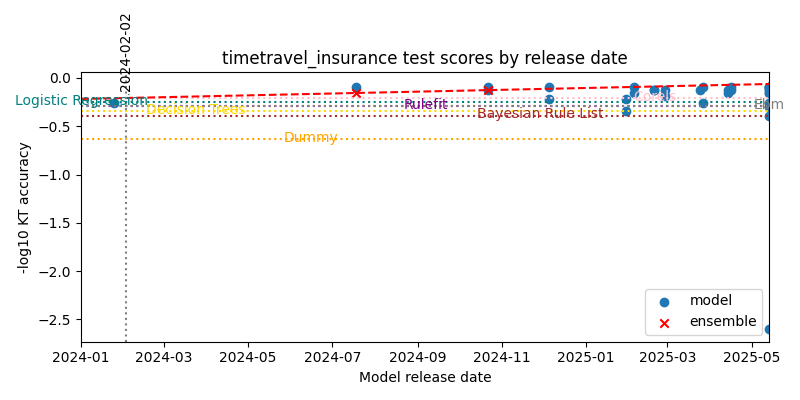}
    \caption{Time Travel}\label{fig:timetravel}
  \end{subfigure}

  \medskip
  \begin{subfigure}[t]{0.49\textwidth}
    \centering
    \includegraphics[width=\linewidth,trim=2mm 2mm 2mm 2mm,clip]{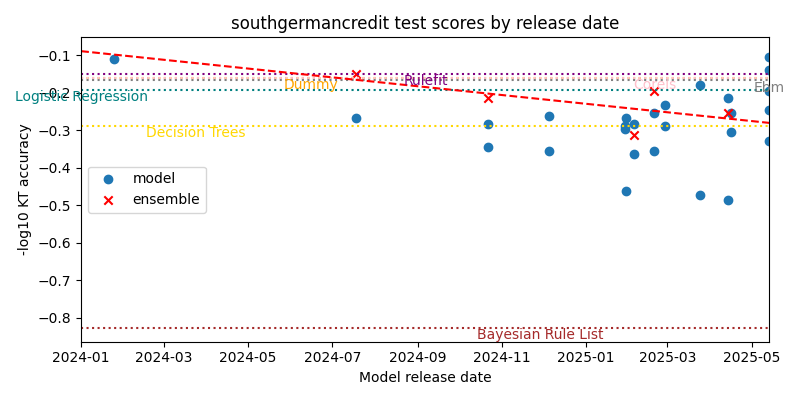}
    \caption{SG Credit}\label{fig:sgcredit}
  \end{subfigure}\hfill
  \begin{subfigure}[t]{0.49\textwidth}
    \centering
    \includegraphics[width=\linewidth,trim=2mm 2mm 2mm 2mm,clip]{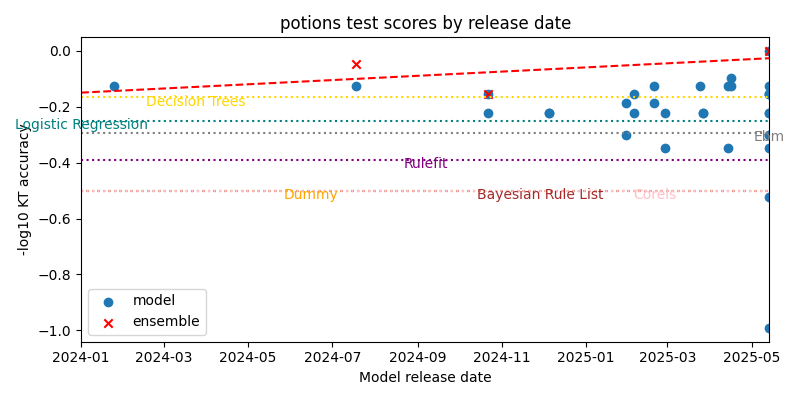}
    \caption{Potions}\label{fig:potions}
  \end{subfigure}

  \caption{Trend over time for ensembles of narrative learners.}
  \label{fig:narrative-trend}
\end{figure*}

\section{Narrative Learning Algorithm}

Narrative Learning is a supervised binary classification algorithm.
The algorithm is shown in Figure~\ref{flow-diagram}.
It is given labelled data, which is split into training, validation and test data.

Two language models are used:

\begin{itemize}
    \item \textbf{Overseer}: Generates classification
      instructions as a natural language prompt. It is only ever shown
      samples from the training data.
    \item \textbf{Underling}: Evaluates each data point independently
      based on the given prompt and returns classifications. It is
      shown samples from the whole dataset.  Since this model will
      potentially do far more work than the Overseer model, ideally it
      will be a smaller, cheaper frontier model. In this experiment we
      used OpenAI's {\tt gpt-4o-mini} and {\tt gpt-4.1-mini} models
      after discovering that {\tt phi-4} was unable to follow overseer
      instructions correctly. 
\end{itemize}

The process is iterative. On the first round, the underlings are given
the prompt ``choose randomly''. The performance metrics (accuracy,
precision, recall, F1 score) are reported to the overseer, along with
a few examples of the correctly-classified and misclassified data
points from the training set. The overseer then refines the prompt
based on what is presented to it, and issues that new prompt back to
the underling(s). The process repeats until the validation scores are
no longer improving. 

The language model is asked to supply its new prompt in a JSON object
where it first provides the narration explaining what needs to be done (giving non-reasoning models an opportunity for some ``reasoning'').  The underlings can also output JSON objects with
this kind of reasoning, but in our experiments we have done no analysis on the
underling reasoning text.

Individual narratives have high variance, so triples of narratives are
ensembled, with a majority vote taken. An open question is
identifying where there is benefit in using larger ensembles: there will
be an additional training cost, but presumably a larger ensemble will
be more accurate; however, there is the practical challenge that there are not
that many frontier-level models.

Every triple combination of individual
narratives is tried, with the successful trio chosen based on their
ensembled success on the validation data.  The accuracy of the ensembled
system is then measured using the test data.

There are two tuneable parameters:
\begin{itemize}
\item The patience of the system.
 the number  of rounds of no validation score improvement before it gives up.
 
\item The number of examples of each
class given to the overseer in each round. Too few examples would be
expected to lead to overfitting and poor predictive performance. Too
many examples and a human would be overwhelmed, and a language model
may exceed its context length --- in both cases making it impossible
to generate a new prompt. 
\end{itemize}

We discuss why we chose to use 3 rounds in Section \ref{parameter-tuning}.
We tried with 3 examples and with 10 examples --- as discussed
in Section
\ref{example-count}.

\section{Experiment Setup}

\begin{table*}
  \begin{tabular}{p{4cm}p{45mm}p{12mm}p{3cm}}
      {\bf Name} & {\bf Transformed into} & {\bf Source} & {\bf Data points} \\
Titanic survival & Medical treatment outcomes & \cite{Titanic2024} & 891 \\
Wisconsin Breast Cancer & Exoplanets & \cite{Wolberg1993} & 569 \\
South German Credit & Coral reef health & \cite{Gromping2019} & 1000 \\
  \end{tabular}
  \caption{Natural datasets used and how they were transformed}
  \label{natural-datasets}
  
\end{table*}

\begin{table*}
  \begin{tabular}{p{35mm}p{20mm}p{3cm}p{10mm}p{10mm}p{10mm}p{10mm}p{10mm}}
    {\bf Name} & {\bf \% noise} & {\bf Data points} & {\bf mean1} & {\bf std1} & {\bf mean2} & {\bf std2} \\
Espionage & 0 & 200 & 70 & 10 & 30 & 8 \\
Timetravel Insurance & 10  & 200 & 12 & 3 & 5 & 2 \\
Magic Potions &  20 & 200 & 40 & 12 & 15 & 5 \\
  \end{tabular}
  \caption{The 2-dimensional synthetic datasets with size, mean and standard deviation of their first and second features}
  \label{synthetic-datasets}
\end{table*}

There is one major challenge in quantifying the effectiveness of
Narrative Learning. Because the agents --- either human beings or
sophisticated language models --- who perform the inferences have
context, training and background knowledge, any common dataset which
has been well studied using traditional machine learning techniques
will be well-known to the agent. For example, it is easy to deduce a
rule for passenger survival on the Titanic. Everyone (human and
non-human alike) knows how unlikely survival was for men and
third-class passengers, and will formulate rules based on that
background knowledge, rather than the knowledge contained in the
examples given.

We address this by transforming each dataset: anonymizing, re-scaling
and relabelling features to prevent reliance on prior
knowledge. Examples of these transformations are given in Appendix~\ref{example-transformations}.  After the experiment is over, we
occasionally un-transform the narratives, to get output like Figure~\ref{titanic-sample}, but the experiment is performed only using the
transformed dataset with no hints available to the language model
about the provenance of the data.

We transformed three datasets as shown in Table~\ref{natural-datasets}

We also created three synthetic datasets with differing levels of
noise. These have never been published and thus it is impossible for the
training process for any frontier model to have seen this data.
These are all two dimensional datasets, with each feature normally distributed
with a given mean and standard deviation, which has a linear classifier
boundary.  Summary statistics are in Table~\ref{synthetic-datasets}

\subsection{Models used}

We created overseers out of a variety of frontier large language models as
shown in Table~\ref{models-used}. We tried using self-hosted models but found
that they could do no better than chance.

\begin{table*}
  \begin{center}
  \begin{tabular}{|p{16mm}|p{10cm}|}
    \hline
        {\bf Vendor}  &  {\bf Model} \\
        \hline
     Anthropic & haiku-3.5, sonnet-3.7, sonnet-4 opus-4  \\
     OpenAI & gpt-4.5-preview, o1, o3, gpt-4o, gpt-4o-mini, gpt-4.1, gpt-3.5 \\
     Google & gemini-2.5-pro-exp, gemini-2.0pro-exp,   gemini-2.0-flash \\
     \hline
  \end{tabular}
\end{center}
\caption{Models used in Narrative Learning experiments}
  \label{models-used}
\end{table*}

\section{Results}

We report our results with both accuracy (Table~\ref{tab:accuracy-scores}) and using the negative log
Krichevsky-Trofimov (KT) accuracy (Table~\ref{tab:negative-log-kt}),
  defined as $S = -\log\left(\frac{c+\frac{1}{2}}{n+1}\right)$ where
  $c$ is the number of correctly classified data points and $n$ is the
  total number of data points evaluated.  Except where otherwise
  noted, we report the score $S$ for the test (held out) dataset.

This choice of KT accuracy as our primary metric needs some justification:

\begin{itemize}
  \item The intuition for this is that we want to report a 100\% successful
classifier on a larger dataset as being ``better'' than a 100\% successful
classifier on a smaller dataset. A KT score of 0.0 would represent perfect
classification accuracy on an infinite dataset.

\item As many of these models (and many of the baselines) were
  producing close-to-perfect results on some datasets, we needed some
  way of providing numbers that meaningfully expressed how close to
  perfection was achieved.

\item It is possible to undo the KT transformation to get accuracy as
  a simple measure, with all the limitations that accuracy has.

\item Since the datasets are mostly quite well balanced, and we do
  compare against a dummy classifier (which predicts the most common class),
  the precision, recall and F1 scores won't provide significantly more
  information than the KT accuracy about the relative success of these
  models compared to other baselines.
\end{itemize}

\newcommand{\colhead}[1]{\multicolumn{1}{c}{\bfseries #1}}
\newcommand{\dsSouthGerman}{\makecell{southgerman\\credit}}
\newcommand{\dsTimeTravel}{\makecell{timetravel\\insurance}}

\begin{table*}
  \centering

\begin{tabularx}{\linewidth}{
  >{\bfseries\raggedright\arraybackslash}X
  *{6}{S[table-format=1.3]}
}

\toprule
\colhead{Model} &
\colhead{espionage} &
\colhead{potions} &
\colhead{\dsSouthGerman} &
\colhead{\dsTimeTravel} &
\colhead{titanic} &
\colhead{wisconsin} \\
    \midrule
    Logistic regression & 0.166 & 0.249 & 0.192 & 0.249 & 0.137 & 0.080 \\
    Decision trees & 0.166 & 0.166 & 0.289 & 0.340 & 0.166 & 0.070 \\
    Dummy & 0.500 & 0.500 & 0.161 & 0.635 & 0.335 & 0.410 \\
    RuleFit & 0.207 & 0.207 & {\bf 0.149} & 0.293 & 0.130 & 0.090 \\
    Bayesian Rule List & 0.081 & 0.500 & 0.827 & 0.389 & 0.481 & 0.142 \\
    CORELS & 0.166 & 0.500 & 0.161 & 0.207 & 0.152 & 0.196 \\
    EBM & 0.166 & 0.293 & 0.167 & 0.293 & 0.116 & 0.070 \\
    \addlinespace
    Most recent successful Narrative Learning ensemble & {\bf 0.071} & {\bf 0.126} & 0.167 & {\bf 0.098} & {\bf 0.106} & {\bf 0.049} \\
    \bottomrule
  \end{tabularx}
  \caption{Negative log KT accuracy ($S$) for baselines compared with the best-performing Narrative Learning ensemble on each dataset. Lower is better.}
  \label{tab:negative-log-kt}
\end{table*}

\begin{table*}
  \centering
\begin{tabularx}{\linewidth}{
  >{\bfseries\raggedright\arraybackslash}X
  *{6}{S[table-format=1.3]}
}

\toprule
\colhead{Model} &
\colhead{espionage} &
\colhead{potions} &
\colhead{\dsSouthGerman} &
\colhead{\dsTimeTravel} &
\colhead{titanic} &
\colhead{wisconsin} \\

    \midrule
    Logistic regression & 0.683 & 0.563 & 0.643 & 0.563 & 0.729 & 0.832 \\
    Decision trees & 0.683 & 0.683 & 0.514 & 0.457 & 0.682 & 0.852 \\
    Dummy & 0.315 & 0.315 & 0.691 & 0.231 & 0.463 & 0.389 \\
    RuleFit & 0.621 & 0.621 & {\bf 0.710} & 0.509 & 0.741 & 0.813 \\
    Bayesian Rule List & 0.832 & 0.315 & 0.149 & 0.408 & 0.330 & 0.721 \\
    CORELS & 0.683 & 0.315 & 0.691 & 0.621 & 0.705 & 0.636 \\
    EBM & 0.683 & 0.509 & 0.681 & 0.509 & 0.765 & 0.852 \\
    Most recent successful Narrative Learning ensemble & {\bf 0.850} & {\bf 0.750}  & 0.681 & {\bf 0.800}  & {\bf 0.784} & {\bf 0.894 } \\
    \bottomrule
  \end{tabularx}
  \caption{Held-out accuracy for baselines and the best-performing Narrative Learning ensemble on each dataset. Higher is better.}
  \label{tab:accuracy-scores}
\end{table*}

Figure~\ref{fig:narrative-trend} shows the trend for $S$ over time for narrative
learning for the six datasets we tested. With the exception of the South German
Credit dataset, the ensemble-triples have been improving as the major LLM
vendors release new models, and have now surpassed all the explainable
language models that we compared against.

The p-value for these trends are statistically significant at $p<0.1$
for most datasets as shown in Figure~\ref{pvalues-and-trends}, but
generally not at $p<0.05$. Note that this would not be solved by
increasing the size of the datasets being processed or taking more
measurements --- this is a trend on the improvements we are seeing
over time as vendors release models. As vendors increasingly take
older models offline, removing public access to them, it is also not
always possible to replicate a long history of model improvements with
new data sets either. The only way to increase the resolving power of
these statistical tests on trends is to start collecting measurements from
as many models as possible today, using data sets that we think might be
difficult for LLMs in the future, and monitor the changes as they happen.
This is unavoidably a longitudinal study.

The potions dataset demonstrates this problem very acutely --- it only took two
improvements to achieve 100\% validation and test accuracy, so no further
improvements can be made, limiting the trend line to two points, about which no $p$ value
can be determined.

The negative trend on the transformed South German Credit dataset is
not significant even at the $p<0.1$ level. We speculate that it may
be performing equivalently to rulefit.

The trend on the transformed Titanic data --- is statistically
significant at $p<0.05$ even when Bonferroni correction is applied.

The trends on the transformed Wisconsin dataset, the Espionage synthetic data and Time Travel
insurance are statistically significant at $p<0.1$ but not when Bonferroni corrected.

\begin{figure}
  \includegraphics[width=0.8\linewidth]{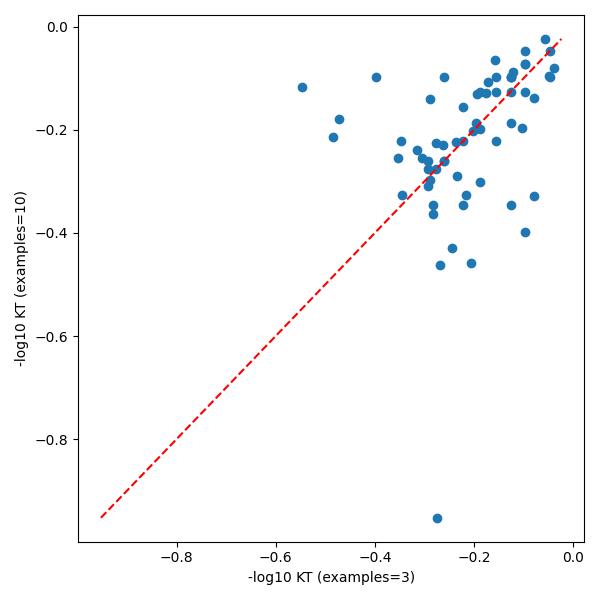}
  \caption{Scatter plot of the accuracy scores from overseers run with 10 examples per round versus being run with 3 examples per round. If 10 examples improved output, the data points would be  highly asymmetric around the central line.}
  \label{example-count-does-not-matter}
\end{figure}

\subsection{Parameter Tuning}
\label{parameter-tuning}

We ran gpt-4o-mini for 100 rounds on each dataset, and in no dataset
did we see an increase in the validation data score that occurred
after an interval longer than 3 rounds. On this basis, we chose
to use a 3-round patience for all other experiments.

\subsection{Example Count}
\label{example-count}

We tested two different example count variants: one where the overseer
was shown 3 examples in each of (true positive, false positive, true negative,
false negative); and a variant where the overseer was shown 10 examples in each
of these classes.

Figure~\ref{example-count-does-not-matter} plots the final accuracy
scores on test scores from overseers run with 10 examples per round
versus being run with 3 examples per round.

If increasing the sample count had a significant effect, we should
expect to see many more points in the upper left --- the area where
the 10-sample accuracy exceeded the 3-sample accuracy. But we do not
--- Figure~\ref{example-count-does-not-matter} is not far from being
symmetric.

The Wilcoxon statistic for whether the 10-example and 3-example
overseers produced different accuracies is 605.00 (p-value: 0.753)
which suggests that indeed any differences are purely chance-related.

If increasing the sample count has an effect, surely it would have been
visible in the jump from 3 to 10. Given that it was not, this hints that
sample counts have little effect on the accuracy of Narrative Learning.

\section{Language complexity}

\begin{figure}
  \begin{center}
    \includegraphics[width=0.9\linewidth]{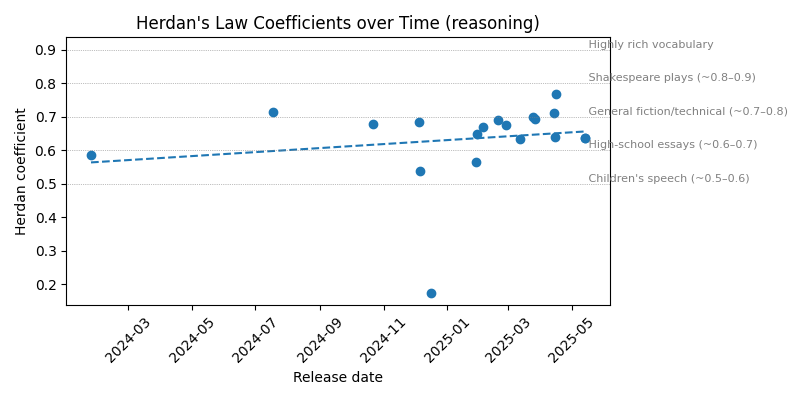} 
    \includegraphics[width=0.9\linewidth]{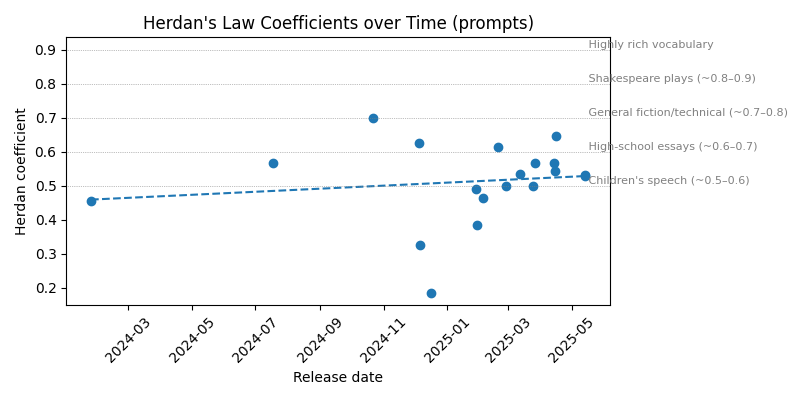}
    \end{center}
    \caption{Narrative Learning is likely to maintain comprehensibility. There is no statistically significant change over time in Herdan's coefficients for the language used in the reasoning over the overseer and the resulting prompts, even while the accuracy of the ensembles improved substantially.      }
  \label{herdan-size-trend}
\end{figure}

A disturbing possibility would be if narrative
learning explanations were getting progressively more complex.
What use would
  a state-of-the-art explainable learning technique  be if the explanations
  are written in language that is so complex and hard to understand
  that they make a Shakespearean play read like a high school essay?

We quantified this using Herdan's
Law\cite{Herdan1960}, which measures lexical diversity,
describing how the number of unique words grows more slowly as texts
get longer.

Shakespeare's works have higher lexical diversity (higher
Herdan's exponent) than the simpler, repetitive vocabulary typically
found in children's books. Classic readability research, such as
Flesch's Reading Ease formulation\cite{Flesch1948}, and more recent
analyses\cite{Tweedie1998} show that more sophisticated language tends
to be harder for readers to process.

Specifically, Herdan's Law states that the number of unique words in a
document grows more slowly than the total number of words, following
the formula $V(n) = k \cdot n^{\beta}$, where $V(n)$ is the vocabulary
size (unique words) after $n$ total words, and $k$ and $\beta$
(Herdan's coefficient) indicates the complexity or richness of the
language used. Higher values of $\beta$ correspond to more varied and
sophisticated vocabulary. For reference, highly repetitive texts like
children's books typically have lower Herdan coefficients (around
$0.5–0.6$), whereas linguistically complex works, such as Shakespeare's
plays, have higher coefficients (around $0.8–0.9$).

The coefficients shown in Figure~\ref{herdan-size-trend} represent the
lexical complexity of language generated by AI models during the
reasoning process and the prompt. Any date on which a new ensemble triple
became the best model for its dataset is plotted, together with the
$\beta$ of that model's prompt and reasoning.

The trend is not statistically significant. For reasoning, the daily
increase is 0.000195 but with a p-value of 0.458. The prompt shows
slightly slower daily growth (0.000146), with a p-value of 0.562.
This does not rule out the possibility that explanations may become
too difficult for us to understand some time in the future, but it
suggests that it is unlikely to be a problem for many years.

\section{Future work}

Investigations into Narrative Learning do not require a particularly sophisticated
understanding of language models or machine learning. It also does not require
an expensive array of GPUs in house, nor a large budget --- we were able to complete
most of these experiments using free tier allocations from the LLM vendors.

The code is available on
github in \url{github.com/solresol/narrative-learning}, and the results are published at \url{https://narrative-learning.symmachus.org/}
but we believe the technique
to be simple enough that it might not need a standard reference implementation.
The data from the experiments
we ran in February through to July 2025 is also available as a postgresql dump.

Together, we expect that this will make Narrative Learning an ideal
undergraduate research project topic. Some possible research questions
(in increasing order of complexity --- the latter ones may be more
appropriate for Masters or PhD students) are:

\begin{itemize}
\item By the time this paper is published, there will be more models
  released. Do the results still hold up with these newer models?
\item Do the results hold up on other datasets?
\item Can we improve the prompt given to the overseer so that it generates
  better prompts for the underling models? That is, can we prompt engineer
  our way to better Narrative Learning results?
\item OpenAI's {\it gpt-4o-mini} and {\it gpt-4.1-mini} models were used for the underling
  model for most experiments, simply because of cost.
  If we substitute another underling model, does that change the results?
\item There are parameters that can be given to language models to control
  inference. Temperature is the most significant of these. How much does
  temperature influence the results?
\item How stable are these models? If we run Narrative Learning multiple
  times using the same model on the same dataset, do we get similar
  results? Are the narratives similar? Are the
  levels of accuracy similar? 
\item Can we measure jumps of insight? Do the overseer models ever
  suddenly get a brand new idea, or do they incrementally improve?
\item Can you systematically evaluate human comprehensibility and interpretability of generated narratives versus classical explainable models? (Potentially via surveys or human evaluations in future work.)
\item How robust is Narrative Learning to noise or adversarial perturbations in data compared to traditional explainable models?
\item It is easy to add as many new obfuscated datasets as desired. What
  properties lead to Narrative Learning being more or less accurate?
\item Nothing in this paper actually needed language models. We could perform this
  experiment using human overseers and underlings. Are language models better at
  hypothesising than humans?
\item How would a multi-class Narrative Learning classifier work?
\item How would a Narrative Learning regressor work?
\item Are the current trends of improvement slowing down? If they are not,
  when would we expect Narrative Learning to be generally
  competitive with the best state-of-the-art models?
\item Is it possible to fine-tune an overseer model to be better at following
  instructions?
\item Are there ways for human experts to work with the overseers to guide them to better hypotheses? How do the results compare to purely AI-driven overseers?
\item If Narrative Learning is applied to scientific datasets with unexplained
  anomalies, does it propose useful narratives the bring insight into that problem?
\end{itemize}

\section{Concluding Observations}

In this paper we have described a new explainable machine learning
algorithm that requires considerably more compute power than other
state-of-the-art explainable models, but typically delivers better
predictive performance, particularly on datasets with low amounts of noise.

We (as data scientists and other users of machine learning) are used
to the idea that training can be probabilistic, but we expect
inference to be deterministic. Narrative Learning does not guarantee
deterministic inference. In early experiments with open source
non-frontier models, underlings would occasionally misinterpret the
narrative. Frontier models eliminated that failure mode, although we
still encountered narratives that were incomplete or impossible to
apply. Even with that non-determinism, the overall inference accuracy
generally exceeded that of deterministic explainable algorithms.

Another unusual aspect to the Narrative Learning algorithm is that ---
unlike logistic regression or the training of neural networks with
gradient descent --- at the level of an individual observation, the
loss function is a binary flag (correct or not). There is no concept
of each iteration of the algorithm getting closer to the truth for one
individual data point. In this regard, it is using the discrete
ultrametric $d(x,y) = 0$ if $x = y$ else $1$. There are other examples
of machine learning algorithms that use the discrete ultrametric (such
as CORELS) but they are somewhat rare.

In this paper we use the terms ``overseer'' and ``underling'', but the
roles could equally be called ``theoretician'' and
``experimentalist'', and the ``narrative'' a ``hypothesis''. The
experimentalist takes a hypothesis and compares its predictions to
held‑out data, rejecting it --- or failing to reject it --- case by
case.  We argue that this alternate terminology reflects how humans do
science. In this regard, we believe that Narrative Learning could also
be used to quantify progress on the Nobel Turing Challenge \cite{Kitano2021NTC}.

\printbibliography

@article{Cox1958,
  author  = {Cox, David R.},
  title   = {The Regression Analysis of Binary Sequences},
  journal = {Journal of the Royal Statistical Society: Series B},
  volume  = {20},
  number  = {2},
  year    = {1958},
  pages   = {215--242}
}

@book{Breiman1984,
  author    = {Breiman, Leo and Friedman, Jerome H. and Olshen, Richard A. and Stone, Charles J.},
  title     = {Classification and Regression Trees},
  publisher = {Wadsworth},
  year      = {1984}
}

@article{Friedman2008,
  author  = {Friedman, Jerome H. and Popescu, Bogdan E.},
  title   = {Predictive Learning via Rule Ensembles},
  journal = {Annals of Applied Statistics},
  volume  = {2},
  number  = {3},
  pages   = {916--954},
  year    = {2008}
}

@article{Letham2015,
  author  = {Letham, Benjamin and Rudin, Cynthia and McCormick, Tyler H. and Madigan, David},
  title   = {Interpretable Classifiers Using Rules and Bayesian Analysis: Building a Better Stroke Prediction Model},
  journal = {Annals of Applied Statistics},
  volume  = {9},
  number  = {3},
  pages   = {1350--1371},
  year    = {2015}
}

@article{Angelino2018,
  author  = {Angelino, Elaine and Larus-Stone, Nicholas and Alabi, Daniel and Seltzer, Michael and Rudin, Cynthia},
  title   = {Learning Certifiably Optimal Rule Lists},
  journal = {Journal of Machine Learning Research},
  volume  = {18},
  number  = {43},
  pages   = {1--78},
  year    = {2018}
}

@inproceedings{Nori2019,
  author    = {Nori, Harsha and Jenkins, Samuel and Koch, Paul and Caruana, Rich},
  title     = {InterpretML: A Unified Framework for Machine Learning Interpretability},
  booktitle = {ICML Workshop on Human Interpretability in Machine Learning},
  year      = {2019}
}

@inproceedings{Ribeiro2016,
  author    = {Ribeiro, Marco Tulio and Singh, Sameer and Guestrin, Carlos},
  title     = {{"Why Should I Trust You?"}: Explaining the Predictions of Any Classifier},
  booktitle = {Proceedings of the 22nd ACM SIGKDD Conference},
  pages     = {1135--1144},
  year      = {2016}
}

@article{Lundberg2017,
  author  = {Lundberg, Scott and Lee, Su-In},
  title   = {A Unified Approach to Interpreting Model Predictions},
  journal = {Advances in Neural Information Processing Systems},
  volume  = {30},
  year    = {2017}
}

@misc{GDPR,
  title        = {Regulation (EU) 2016/679: General Data Protection Regulation},
  howpublished = {Official Journal of the European Union},
  year         = {2016},
  note         = {Article 22 \& Recital 71}
}

@misc{PIPL,
  title        = {Personal Information Protection Law of the People's Republic of China},
  howpublished = {Order No. 91},
  year         = {2021},
  note         = {Article 24 on automated decision-making}
}

@misc{Titanic2024,
  title        = {Titanic --- Machine Learning from Disaster Dataset},
  howpublished = {Kaggle},
  year         = {2024},
  note         = {Accessed 20 July 2025}
}

@dataset{Wolberg1993,
  author  = {Wolberg, William and Mangasarian, Olvi and Street, William},
  title   = {Breast Cancer Wisconsin (Diagnostic) Data Set},
  howpublished = {UCI Machine Learning Repository},
  year    = {1993}
}

@techreport{Gromping2019,
  author     = {Gr\"{o}mping, Ulrike},
  title      = {South German Credit Data: Correcting a Widely Used Data Set},
  institution = {Beuth University of Applied Sciences Berlin},
  number     = {Report 4/2019},
  year       = {2019}
}

@book{Herdan1960,
  author    = {Herdan, Gustav},
  title     = {Type--Token Mathematics},
  publisher = {Mouton},
  year      = {1960}
}

@article{Flesch1948,
  author  = {Flesch, Rudolf},
  title   = {A New Readability Yardstick},
  journal = {Journal of Applied Psychology},
  year    = {1948},
  volume  = {32},
  number  = {3},
  pages   = {221--233},
  doi     = {10.1037/h0057532}
}

@article{Tweedie1998,
  author  = {Tweedie, Fiona J. and Baayen, R. Harald},
  title   = {How Variable May a Constant Be? Measures of Lexical Richness in Perspective},
  journal = {Computers and the Humanities},
  volume  = {32},
  pages   = {323--352},
  year    = {1998}
}

@article{Kitano2021NTC,
  author  = {Kitano, Hiroaki},
  title   = {Nobel Turing Challenge: Creating the Engine for Scientific Discovery},
  journal = {npj Systems Biology and Applications},
  volume  = {7},
  number  = {29},
  doi     = {10.1038/s41540-021-00189-3},
  year    = {2021}
}

\onecolumn
\appendix

\section{Example Transformations}\label{example-transformations}

The following sections contain the text of some of the conversion instructions given to the
obfuscation tool.

Note that these obfuscations prompts themselves were generally AI-created, with
only minor tweaks done by hand.


\subsection{Titanic}

\begin{quote}
  {
  \sf
This is the famous Titanic dataset. What I want to do is convert this
to a dataset that looks like it is for a medical treatment. I want to
mask everything but keep the details the same.

\begin{enumerate}
\item Survived changes from (0,1) to (Success, Failure)
\item PassengerId should be Patient\_ID
\item Pclass should be renamed to "Histogen\_Complex" and change (1,2,3) to (Beta, Omicron, Delta)
\item Name should be dropped
\item Sex should be inverted.
\item Age should become "Treatment\_Months", and be 3 times whatever the age was. Where the Age is null, impute a mean.
\item SibSp should become "Genetic\_Class\_A\_Matches" and be one more than SibSp.
\item Parch should be "Genetic\_Class\_B\_Matches" and be one more than Parch
\item Fare should become "TcQ\_mass" and be 1000 times fare.
\item Drop Cabin
\item Change Embarked to "Cohort" with (S,C,Q) -> (Melbourne, Delhi, Lisbon)
\end{enumerate}
}
\end{quote}

\subsection{Wisconsin Breast Cancer}

\begin{quote}
  {\sf
Let's completely disguise this dataset. Turn it into a space science
dataset, making it look like data from an astronomical survey of
exoplanets. Here's the plan:

Column Renaming (Space Science Theme)

\begin{longtable}{p{3cm}p{5cm}p{5cm}}
\toprule
{\bf Original Column} & {\bf New Column (Space Science)} & {\bf Interpretation}
\endhead
patient\_id & exoplanet\_id & Unique identifier for each exoplanet \\
mean radius & mean\_orbital\_radius & Average distance of the exoplanet
from its star \\
mean texture & mean\_surface\_roughness & Surface irregularities of the
exoplanet \\
mean perimeter & mean\_magnetosphere\_extent & Strength of the planet's
magnetic field boundary \\
mean area & mean\_atmospheric\_depth & Average depth of the planet's
atmosphere \\
mean smoothness & mean\_tidal\_distortion & Degree to which the planet's
shape is deformed by tidal forces \\
mean compactness & mean\_core\_density & Density of the planet's core \\
mean concavity & mean\_ring\_system\_complexity & Complexity of rings
around the planet (if any) \\
mean concave points & mean\_impact\_crater\_count & Number of impact
craters detected on the surface \\
mean symmetry & mean\_axial\_symmetry & How symmetrical the planet
appears in infrared imaging \\
mean fractal dimension & mean\_cloud\_turbulence & Measure of the
complexity of cloud formations in the planet's atmosphere \\
radius error & orbital\_radius\_error & Measurement uncertainty in the
orbital radius \\
texture error & surface\_roughness\_error & Measurement uncertainty in
surface roughness \\
perimeter error & magnetosphere\_extent\_error & Error in magnetic
boundary estimates \\
area error & atmospheric\_depth\_error & Error in atmospheric depth
measurements \\
smoothness error & tidal\_distortion\_error & Uncertainty in tidal
distortion measurement \\
compactness error & core\_density\_error & Error in core density
estimation \\
concavity error & ring\_system\_complexity\_error & Error in measuring
ring complexity \\
concave points error & impact\_crater\_count\_error & Uncertainty in
crater count \\
symmetry error & axial\_symmetry\_error & Uncertainty in axial symmetry
measurement \\
fractal dimension error & cloud\_turbulence\_error & Uncertainty in
cloud turbulence estimation \\
worst radius & max\_orbital\_radius & Largest measured orbital radius \\
worst texture & max\_surface\_roughness & Highest recorded surface
roughness \\
worst perimeter & max\_magnetosphere\_extent & Largest estimated
magnetosphere size \\
worst area & max\_atmospheric\_depth & Thickest atmospheric
measurement \\
worst smoothness & max\_tidal\_distortion & Maximum recorded tidal
deformation \\
worst compactness & max\_core\_density & Highest estimated core
density \\
worst concavity & max\_ring\_system\_complexity & Most complex ring
system observed \\
worst concave points & max\_impact\_crater\_count & Highest recorded
impact crater count \\
worst symmetry & max\_axial\_symmetry & Highest detected axial
symmetry \\
worst fractal dimension & max\_cloud\_turbulence & Most chaotic cloud
formations detected \\
target & occupying\_species & Fylaran or Qtharri \\
\end{longtable}

Numerical Transformations

To ensure the LLM can't reverse-engineer the dataset, transform the
numbers while preserving the underlying relationships.
\begin{itemize}
  \item For
size-related metrics (radius, area, perimeter, etc.): Apply a
logarithmic transformation (log(x + 1)) to smooth out large variations.
\item  For texture/symmetry measures: Convert them to ranks within the
  dataset to remove absolute values while preserving order.

\item For errors
and variation measures: Apply a square root transformation (sqrt(x)) to
make them less extreme while keeping relative differences.

\item For
compactness and density-like measures: Apply a reciprocal transformation
(1 / (x + 1)) to flip the scale.

\item For binary classification (target $\rightarrow$
occupying\_species): Where it had the word malignant, make that
``Fylaran'' and change the word ``benign'' to ``Qtharri''.
\end{itemize}

}
\end{quote}

\section{Training prompt}

The overseer model is shown this prompt
(followed by examples of true positives, true negatives, false positives
and false negatives) in each round of training.

{\tt

You are part of a program that is trying to learn inference rules on
this dataset. At each round, a prompt is shown to an LLM together with
one row of data at a time. It then attempts to predict the outcome based on the
rules in the prompt. This process works well if the prompt has very
explicit and clear rules: aim for unambiguous thresholds for values,
clear criteria for labels and careful wording.

We would like to improve the prompt that is being used.

Please create a new prompt that will reduce the number of false
positives and false negatives in this dataset. You can see the
prompt(s) that have been used previously, and how effective they
were. There are also some examples of where those prompt(s) did and
didn't work.

Remember: you need to create rules. Don't just waffle about what
changes need to happen. Look at the examples where the previous prediction
system got it wrong, and try to come up with at least one new rule that
would handle one of those situations correctly.

}

\subsection{Rules generated}
\label{rules-generated}

The following sections show the narrative models created for the
medically-transformed Titanic dataset, which have then been
mechanically untransformed so that they refer to the columns
of the original Titanic dataset.

\subsection{claude-3-5-haiku-20241022, with 10 examples in each round}

\begin{verbatim}
Comprehensive Survival Probabilistic Prediction Model v4.0:

1. Advanced Fare Evaluation (40% weight):
   a) Hyper-Adaptive Variant-Specific Scoring:
      - Variant Transformation Strategies:
        * Pclass 2: 
          - Optimal Range: 7 - 48
          - Quadratic scoring with asymmetric uncertainty zones
          - Sharp non-linear penalties for extreme deviations
        
        * Pclass 3: 
          - Optimal Range: 9 - 35 (Refined Lower Bound)
          - Adaptive sigmoid function with enhanced sensitivity
          - Explicit low-value success probability modeling
          - Contextual edge case handling for short-duration treatments
        
        * Pclass 1:
          - Optimal Range: 22 - 115
          - Polynomial scoring with machine learning refinement
          - Adaptive margin sensitivity
          - Interaction-aware scoring mechanism

   b) Probabilistic Transformation:
      - Machine learning-inspired continuous probability generation
      - Explicit local and global uncertainty quantification
      - Variant-specific interaction modeling

2. Enhanced Age Probability (25% weight):
   a) Dynamic Duration Modeling with Adaptive Learning:
      - Variant-Specific Intelligent Thresholds:
        * Pclass 2: 5-32 years (peak: 15-25 years)
        * Pclass 3: 3-18 years (peak: 8-14 years) - Refined for shorter treatments
        * Pclass 1: 13-52 years (peak: 23-41 years)
      
      - Advanced Probabilistic Success Curve:
        * Adaptive logistic function with machine learning refinement
        * Variant-specific probability transition modeling
        * Explicit short-duration success probability for Pclass 3
        * Interaction-weighted duration scoring

3. Sophisticated Family Match Evaluation (20% weight):
   a) Advanced Matching Strategy with Interaction Intelligence:
      - SibSp Matches: Primary Predictive Power
      - Parch Matches: Complementary Validation
      - Enhanced Cross-Class Interaction Modeling
   
   b) Nuanced Match Scoring Algorithm:
      - 0 matches: Near-zero success probability with explicit uncertainty
      - 1 match: Probabilistic base with interaction potential
      - 2 matches: Significant probability boost with adaptive weighting
      - 3+ matches: Carefully modeled diminishing returns
      - Dynamic match distribution bonus
      - Explicit cross-class interaction scoring
      - Variant-specific match interaction refinement

4. Contextual Probabilistic Modifiers (15% weight):
   a) Refined Embarkation Influence with Adaptive Learning:
      - Southampton: Dynamic positive modifier (+10% to +15%)
      - Cherbourg: Adaptive neutral modifier (-2% to +5%)
      - Queenstown: Contextual negative modifier (-5% to -12%)
      - Enhanced interaction-aware factor weighting
   
   b) Advanced Sex-Based Nuancing:
      - Probabilistic adjustment curves with machine learning refinement
      - Explicit interaction with genetic, duration, and group factors
      - Uncertainty-aware, dynamically weighted modifications

5. Comprehensive Probabilistic Integration:
   a) Multi-Factor Interaction Model with Adaptive Intelligence:
      - Non-linear multiplicative integration
      - Machine learning-inspired interaction term modeling
      - Preserving individual factor characteristics
      - Dynamic factor importance adjustment
   
   b) Advanced Uncertainty Quantification:
      - 95% Confidence Interval Generation with Adaptive Boundaries
      - Granular Factor-Level Contribution Analysis
      - Probabilistic Output: Continuous [0, 1] scale
      - Explicit Uncertainty Magnitude Indicator
      - Learning-based confidence refinement

Prediction Output:
- Survival Probability (0-1 continuous)
- 95% Confidence Interval with Adaptive Boundaries
- Key Determining Factors with Interaction Insights
- Uncertainty Magnitude and Trend
- Detailed Factor Contributions and Interactions

Interpretation Guidelines:
- Probability > 0.75: High Likelihood of Survival
- Probability 0.50-0.75: Moderate Uncertainty with Detailed Insights
- Probability < 0.50: Low Survival Likelihood

Critical Caveat: Advanced probabilistic guidance with transparent uncertainty communication,
acknowledging individual passenger variability and complex interactions.

Recommendation: Utilize comprehensive, adaptive probabilistic prediction with explicit
interaction modeling and dynamic uncertainty quantification, with special attention to
Pclass 3 short-duration treatments.
\end{verbatim}

\subsection{claude-3-5-haiku-20241022, 3 examples per round}
\begin{verbatim}
Predict Survived = 1 if ANY of these criteria are met:

1. Pclass = 2 AND Fare > 20
OR
2. ALL of these conditions are met:
   - Fare > 12
   - (SibSp + Parch) >= 1
   - Age >= 2.0
OR
3. ALL of these conditions are met:
   - Age > 8.0
   - Fare > 15
   - SibSp >= 1

Otherwise, predict Survived = 0.

Note: These criteria are particularly strict to reduce false positives, which were a major
issue in the previous results.
\end{verbatim}

\subsection{OpenAI gpt-4o with 10 examples}
\begin{verbatim}
You are tasked with predicting the survival or death of passengers based on their data. Follow
these explicit rules to improve the accuracy of your predictions: 

1. **Passenger Class**:
   - Results tend to vary, but common survival was seen in '1st' and '3rd' passenger classes.

2. **Sex**:
   - While varied, gender alone should not weigh heavily unless interfaced with other factors.

3. **Age**:
   - Be cautious: older age is not necessarily indicative of survival or death. 
   - Rule: Consider more closely when age is extreme (either very young <3 years
     or very old >33.33 years).

4. **Family Size**:
   - **SibSp >= 0** and **Parch >= 0**: This often correlates with survival, especially
     when both have at least one family member.
   - Greater family size could weigh more heavily towards prediction survival.

5. **Fare**:
   - Avoid predicting survival if fare is too low. Prioritize when fare is above 20.0.

6. **Embarked**:
   - Consider separately for different embarkation points, as place might affect survival.
     Currently more examples come from Southampton.

Proceed to predict outcomes adhering strictly to these guidelines, carefully balancing each
factor's contribution to make a reasoned decision. Avoid bias by over-relying on any singular
 attribute and instead synthesize information to match explicit patterns outlined here. Aim
for minimizing both false positives and negatives.
\end{verbatim}

\subsection{OpenAI gpt-4o}
\begin{verbatim}
To enhance prediction accuracy and reduce false positives and negatives in survival outcomes,
follow these detailed criteria:

1. Passenger Class Criteria:
   - '2nd' class passengers have the highest survival probability; prediction rules can be
slightly relaxed for '3rd' class, while '1st' class requires at least one exceptional
condition to be met (e.g., high family connections or ideal fare).

2. Family Connection Criteria:
   - Survival is more likely with at least one sibling/spouse and one parent/child aboard.
Alternatively, a total of two or more family connections (sibling/spouse and/or parent/child)
indicates predicted survival.

3. Age Guidelines:
   - Survival is predicted for ages strictly between 10 and 26. For ages between 27 and 40,
survival is conditional on strong additional family support or exceptional fare.

4. Fare Specifics:
   - Predict survival if fare is precisely within the range of 16 to 24.5. For values between
24.501 and 26.25, survival is suggested only with additional family support.

5. Long Duration Critical Caution:
   - Ages exceeding 40 should generally be predicted as non-survival unless offset by two
of the following: exceptional family connections (two or more combined connections), idealized
fare, or embarkation-based exception proven to show survival historically.

Integrate assessments holistically, acknowledging interaction effects, to minimize prediction
errors and optimize both precision and recall. Apply rigorous logic to ensure all criteria are
met satisfactorily before classifying a passenger as likely to survive.
\end{verbatim}

\end{document}